\documentclass[11pt]{article}

\usepackage{amsfonts}
\usepackage{amssymb}
\usepackage{amsmath} 
\usepackage{amsthm}
\usepackage{graphicx}

\newcommand {\Xcal}{\mathcal{X}}
\newcommand{\RR}{\mathbb{R}}    
\newcommand{\NN}{\mathbb{N}}    
\newcommand{\Hcal}{\mathcal{H}} 
\newcommand {\br}[1]{\left(#1\right)}
\newcommand {\inpH}[2]{\left\langle #1 \right\rangle_{#2}}
\newcommand {\cbr}[1]{\left\{#1 \right\}}

\newtheorem{defi}{Definition}
\newtheorem{thm}{Theorem}
\newtheorem{rem}{Remark}
\newtheorem{lem}{Lemma}

\begin{document}

\title{The optimal assignment kernel is not positive definite}
\author{Jean-Philippe Vert\\
Centre for Computational Biology\\
Mines ParisTech\\
\texttt{Jean-Philippe.Vert@mines.org}}
\date{\today}
\maketitle
\begin{abstract}
We prove that the optimal assignment kernel, proposed recently as an attempt to embed labeled graphs and more generally tuples of basic data to a Hilbert space, is in fact not always positive definite.
\end{abstract}

\section{Introduction}
Let $\Xcal$ be a set, and $k:\Xcal\times\Xcal\rightarrow\RR$ a symmetric function that satisfies, for any $n\in\NN$ and any $\br{a_{1},\ldots,a_{n}}\in\RR^n$ and $\br{x_{1},\ldots,x_{n}}\in\Xcal^n$:
$$
\sum_{i=1}^n\sum_{j=1}^n a_{i}a_{j} k(x_{i},x_{j}) \geq 0\,.
$$
Such a function is called a \emph{positive definite kernel} on $\Xcal$. A famous result by \cite{Aronszajn1950Theory} states the equivalence between the definition of a positive kernel and the embedding of $\Xcal$ in a Hilbert space, in the sense that $k$ is a positive definite kernel on $\Xcal$ if and only if there exists a Hilbert space $\Hcal$ with inner product $\inpH{\cdot,\cdot}{\Hcal}$ and a mapping $\Phi:\Xcal\rightarrow\Hcal$ such that, for any $x,x'\in\Xcal$, it holds that:
\begin{equation}\label{eq:inp}
k(x,x') = \inpH{\Phi(x),\Phi(x')}{\Hcal}\,.
\end{equation}

The construction of positive definite kernels on various sets $\Xcal$ has recently received a lot of attention in statistics and machine learning, because they allow the use of a variety of algorithms for pattern recognition, regression of outlier detection for sets of points in $\Xcal$ \cite{Vapnik1998Statistical,Scholkopf2002Learning}. These algorithms, collectively referred to as \emph{kernel methods}, can be thought of as multivariate linear methods that can be performed on the Hilbert space implicitly defined by any positive definite kernel $k$ through (\ref{eq:inp}), because they only access data through inner products, hence through the kernel. This ``kernel trick'' allows, for example, to perform supervised classification or regression on strings or graphs with state-of-the-art statistical methods as soon as a positive definite kernel for strings or graphs is defined. Unsurprisingly, this has triggered a lot of activity focused on the design of specific positive definite kernels for specific data, such as strings and graphs for applications in bioinformatics in natural language processing \cite{Shawe-Taylor2004Kernel}.

Motivated by applications in computational chemistry, \cite{Frohlich2005Optimal} proposed recently a kernel for labeled graphs, and more generally for structured data that can be decomposed into subparts. The kernel, called \emph{optimal assignment kernel}, measures the similarity between two data points by performing an optimal matching between the subparts of both points. It translates a natural notion of similarity between graphs, and can be efficiently computed with the Hungarian algorithm. However, we show below that it is in general not positive definite, which suggests that special care may be needed before using it with kernel methods.

It should be pointed out that not being positive definite is not necessarily a big issue for the use of this kernel in practice. First, it may in fact be positive definite when restricted to a particular set of data used in a practical experiment. Second, other non positive definite kernels, such as the sigmoid kernel, have been shown to be very useful and efficient in combination with kernel methods. Third, practitioners of kernel methods have developed a variety of strategies to limit the possible dysfunction of kernel methods when non positive definite kernels are used, such as projecting the Gram matrix of pairwise kernel values on the set of positive semidefinite matrices before processing it. The good results reported on several chemoinformatics benchmark in \cite{Frohlich2005Optimal} indeed confirm the usefulness of the method. Hence our message in this note is certainly not to criticize the use of the optimal assignment kernel in the context of kernel methods. Instead we wish to warn that in some cases, negative eigenvalues may appear in the Gram matrix and specific care may be needed, and simultaneously to contribute to the limitation of error propagation in the scientific litterature.

\section{Main result}
Let us first define formally the optimal assignment kernel of \cite{Frohlich2005Optimal}. We assume given a set $\Xcal'$, endowed with a positive definite kernel $k_{1}$ that takes only nonnegative values. The objects we consider are tuples of elements in $\Xcal'$, i.e., an object $x$ decomposes as $x=\br{x_{1},\ldots,x_{n}}$, where $n$ is the length of the tuple $x$, denoted $|x|$, and $x_{1},\ldots,x_{n}\in\Xcal'$. We note $\Xcal$ the set of all tuples of elements in $\Xcal'$. Let $S_{n}$ be the symmetric group, i.e., the set of permutations of $n$ elements. We now recall the kernel on $\Xcal$ proposed in \cite{Frohlich2005Optimal}:
\begin{defi}
The \emph{optimal assignment kernel} $k_{A}:\Xcal\times\Xcal\rightarrow \RR$ is defined, for any $x,y\in\Xcal$, by:
$$
k_{A}(x,y) =
\begin{cases}
\max_{\pi\in S_{|x|}} \sum_{i=1}^{|x|} k_{1}(x_{i},y_{\pi(i)}) & \text{ if }|y|\geq|x|\,,\\
\max_{\pi\in S_{|y|}} \sum_{i=1}^{|y|} k_{1}(x_{\pi(i)},y_{i}) & \text{ otherwise.}
\end{cases}
$$ 
\end{defi}
We can now state our main theorem.
\begin{thm}\label{thm1}
The optimal alignment kernel is not always positive definite.
\end{thm}

Before proving this results we can make a few comments.
\begin{rem}
The meaning of the statement ``not always'' in Theorem \ref{thm1} is that there exist choices of $\Xcal'$ and $k_{1}$ such that the optimal assignment kernel is positive definite, while there also exist choices for which it is not positive definite.
\end{rem}
\begin{rem}
Theorem \ref{thm1} contradicts Theorem 2.3 in \cite{Frohlich2005Optimal}, which claims that the optimal assignment kernel is always positive definite. The proof of Theorem 2.3 in \cite{Frohlich2005Optimal}, however, contains the following error. Using the notations of \cite{Frohlich2005Optimal}, the author define in the course of their proof the values $A:=2\sum_{j=1}^n v_{n+1} v_{j} K_{n+1,j}$ and $B:=\sum_{j=1}^n v_{n+1}^2 K_{n+1,n+1} + v_{j}^2 K_{jj}$. They show that $A\leq B$, on the one hand, and that $A<0$, on the other hand. From this they conclude that $B<0$, which is obvioulsy not a valid logical conclusion. 
\end{rem}

In order to prove Theorem \ref{thm1}, we now provide an example of $\br{\Xcal',k_{1}}$ pair that leads to a positive definite optimal assignment kernel, and another example that leads to the opposite conclusion.
\begin{lem}
Let $\Xcal'=\cbr{1}$ be a singleton, and $k_{1}(1,1)=1$. Then the optimal assignment kernel is positive definite.
\end{lem}
\begin{proof}
When $\Xcal'=\cbr{1}$, the tuples are simply repeats of the unique element, hence each element $x=\br{1,\ldots,1} \in\Xcal$ is uniquely defined by its length $|x|\in\NN$. The optimal assignment kernel is then given by:
$$
k_{A}(x,y) = \min\br{|x|,|y|}\,.
$$
The function $\min\br{a,b}$ is known to be positive definite on $\NN$, therefore $k$ is a valid kernel on $\Xcal$.
\end{proof}
\begin{lem}\label{lem2}
Let $\Xcal' = \RR^2$ and $k_{1}(x,y) = \exp\br{-\gamma||x-y||^2}$, for $x,y\in\RR^2$ and $\gamma>0$. Then the optimal assignment kernel is not positive definite.
\end{lem}
\begin{proof}
The function $k_{1}$ defined in Lemma \ref{lem2} is the well-known Gaussian radial basis function kernel, which is known to be positive definite and only takes nonnegative values, hence it satisfies all hypothesis needed in the definition of the optimal assignment kernel. In order to show that the latter is not positive definite, we exhibit a set of points in $\Xcal$ that can not be embedded in a Hilbert space through (\ref{eq:inp}). For this let us start with four points that form a square in $\Xcal'$, e.g., $A=(0,0), B=(1,0), C=(1,1)$ and $D=(0,1)$ (Figure \ref{fig:square}).
\begin{figure}[htbp]
\begin{center}
\includegraphics[width=4cm]{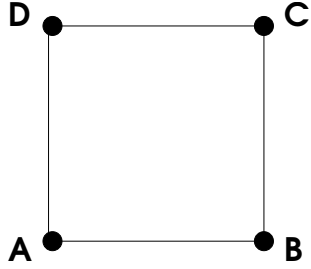}
\caption{Four points in $\Xcal'=\RR^2$, endowed with the positive definite kernel $k_{1}(x,y) = \exp\br{-\gamma||x-y||^2}$.}
\label{fig:square}
\end{center}
\end{figure}
Denoting $a:=\exp(-\gamma)$, we directly obtain from the definition of $k_{1}$ that:
\begin{equation*}
\begin{cases}
k_{1}(A,A)=k_{1}(B,B)=k_{1}(C,C)=k_{1}(D,D)&=1\,,\\
k_{1}(A,B) = k_{1}(B,C) = k_{1}(C,D) = k_{1}(D,A) &= a\,,\\
k_{1}(A,C) = k_{1}(B,D) &=a^2\,.
\end{cases}
\end{equation*}

In the space $\Xcal$ of tuples, let us now consider the six $2$-tuples obtained by taking all pairs of distinct points: $AB, AC, AD, BC, BD, CD$. Using the definition of the optimal assignment kernel $k(uv,wt)=\max(k_{1}(u,w) + k_{1}(v,t) , k_{1}(u,t) + k_{1}(v,w))$ for $u,v,w,t\in \cbr{A,B,C,D}$, we easily obtain:
$$
\begin{cases}
k(AB,AB)=k(AC,AC)=k(AD,AD)=k(BC,BC)\\
\hspace{5cm}=k(BD,BD)=k(CD,CD)=2\,,\\
k(AB,AC)=k(AB,BD)=k(BC,BD)=k(BC,AC)\\
\hspace{1cm}=k(CD,AC)=k(CD,BD)=k(AD,AC)=k(AD,BD)=1+a\,,\\
k(AB,BC)=k(BC,CD)=(CD,AD)=k(AB,AD)=1+a^2\,,\\
k(AB,CD)=k(AD,BC)=k(AC,BD)=2a\,.
\end{cases}
$$
If $k$ was positive definite, then these six $2$-tuples could be embedded to a Hilbert space $\Hcal$ by a mapping $\Phi:\Xcal\rightarrow\Hcal$ satisfying (\ref{eq:inp}). Let us show that this is impossible. Let $d(x,y)=||\Phi(x)-\Phi(y)||_{\Hcal}$ be the Hilbert distance between two points $x,y\in\Xcal$ after their embedding in $\Hcal$. It can be computed from the kernel values by the classical equality:
$$
d(x,y)^2 = k(x,x)+k(y,y)-2k(x,y)\,.
$$
We first observe that $d(AB,AC)^2=d(AC,CD)^2=2-2a$, and $d(AB,CD)^2=4-4a$. Therefore,
$$
d(AB,CD)^2 = d(AB,AC)^2 + d(AC,CD)^2\,,
$$
from which we conclude that $(AB,AC,CD)$ form a half-square, with hypotenuse $(AB,CD)$. A similar computation shows that $(AB,BD,CD)$ is also a half-square with hypotenuse $(AB,CD)$. Moreover,
$$
d(AC,BD) = 4-4a = d(AB,CD)\,,
$$
which shows that the four points $(AB,AC,CD,BD)$ are in fact coplanar and form a square. The same computation when $AB$ and $CD$ are respectively replaced by $AD$ and $BC$ shows that the four points $(AD,AC,BC,BD)$ are also coplanar and also form a square. Hence all six points can be embedded in $3$ dimensions, and the points $(AB,AD,CD,BC)$ are themselves coplanar and must form a rectangle on the plane equidistant from $AC$ and $BD$ (Figure \ref{fig2}).
\begin{figure}[htbp]
\begin{center}
\includegraphics[width=5cm]{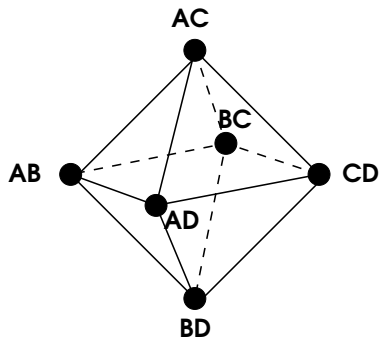}
\caption{The necessary configuration of the six $2$-tuples if embedding with the optimal assignment kernel was possible.}
\label{fig2}
\end{center}
\end{figure}
The edges of this rectangle have all the same length $d(AB,BC)^2 = d(BC,CD)^2 = d(CD,AD)^2 = d(AD,AB)^2 = 2-2a^2$ and is therefore a square, whose hypotenuse $(AB,CD)$ should have a length $\sqrt{4-4a^2}$. However a direct computation gives $d(AB,CD)=\sqrt{4-4a}$, which provides a contradiction since $0<a<1$. Hence the six points can not be embedded into a Hilbert space with $k$ as inner product, which shows that $k$ is not positive definite on $\Xcal$.
\end{proof}
\bibliographystyle{plain}

\end{document}